\documentclass[sigconf,nonacm]{acmart}

\AtBeginDocument{%
  }

\copyrightyear{2026}
\acmYear{2026}

\begin{document}

\title{Generalizable Operating Room Expert with Multimodal Enhancement}

\author{Peiqi He}
\authornote{Both authors contributed equally to this research.}
\email{hepeiqi@hnu.edu.cn}
\orcid{0009-0005-1690-7402}
\affiliation{%
  \institution{Hunan University}
  \city{Changsha}
  \state{Hunan}
  \country{China}
}
\author{Zhenhao Zhang}
\authornotemark[1]
\authornote{Zhenhao Zhang led project administration and coordinated the overall planning, implementation, and evaluation of this work.}
\email{zhangzhh2024@shanghaitech.edu.cn}
\affiliation{%
  \institution{ShanghaiTech University}
  \city{Shanghai}
  \country{China}
}
\author{Yixiang Zhang}
\email{zhangyixiang@hnu.edu.cn}
\affiliation{%
  \institution{Hunan University}
  \city{Changsha}
  \state{Hunan}
  \country{China}
}
\author{Jiaxin Liu}
\email{liujx2024@shanghaitech.edu.cn}
\affiliation{
  \institution{ShanghaiTech University}
  \city{Shanghai}
  \country{China}
}

\author{Xiongjun Zhao}
\correspondingauthor
\email{xiongjunzhao@hnu.edu.cn}
\affiliation{%
  \institution{Hunan University}
  \city{Changsha}
  \state{Hunan}
  \country{China}
}
\author{Shaoliang Peng}
\correspondingauthor
\email{slpeng@hnu.edu.cn}
\affiliation{%
  \institution{Hunan University}
  \city{Changsha}
  \state{Hunan}
  \country{China}
}

\renewcommand{\shortauthors}{Peiqi He et al.}

\begin{abstract}
Precise spatial modeling in the operating room (OR) is essential for intraoperative awareness, hazard avoidance, and surgical decision-making. Although existing approaches exploit multimodal data to learn spatial relationships, many depend on sensing modalities that are difficult to deploy in real clinical environments and remain limited in explicit 3D reasoning under constrained sensing conditions. Meanwhile, models trained primarily on readily available 2D data often fail to capture the fine-grained geometric and semantic structure of complex OR scenes. To address these limitations, we introduce \textbf{OR-Expert}, a large vision-language framework for 3D spatial reasoning with RGB-only inference. OR-Expert internally derives depth, panoptic segmentation, and point-cloud cues from RGB images and encodes them as structured spatial representations. Its Spatial-Enhanced Feature Fusion Block aligns these pseudo-modalities with RGB and textual features in a shared token space, enabling joint semantic, geometric, and language reasoning. The unified end-to-end MLLM therefore supports detailed spatial understanding without requiring external depth, segmentation, or point-cloud sensors, or additional expert annotations at inference time. Experiments on multiple operating-room benchmarks demonstrate that OR-Expert achieves state-of-the-art performance and generalizes effectively to unseen surgical scenes and downstream spatial reasoning tasks.
\end{abstract}


\maketitle

\section{Introduction}

Modern operating rooms (ORs) are complex, high-stakes environments in which accurate spatial awareness is essential for surgical safety, workflow coordination, and intraoperative decision-making. Clinicians and intelligent systems must continuously interpret dynamic spatial relationships, including the positions of medical staff, patients, surgical instruments, and surrounding equipment. However, robust 3D spatial reasoning remains challenging under the constrained sensing conditions of real-world clinical environments. In many hospitals, surgical data are primarily captured using standard monocular RGB cameras, whereas depth sensors, multi-view systems, and dense semantic annotations are often unavailable because of equipment costs, infrastructure constraints, and potential disruption to established clinical workflows. Consequently, models must infer geometric and semantic structure from visually complex scenes characterized by occlusion, reflective surfaces, dense object interactions, and frequent viewpoint changes.

\begin{figure}[!h]
    \centering
    \includegraphics[width=\linewidth]{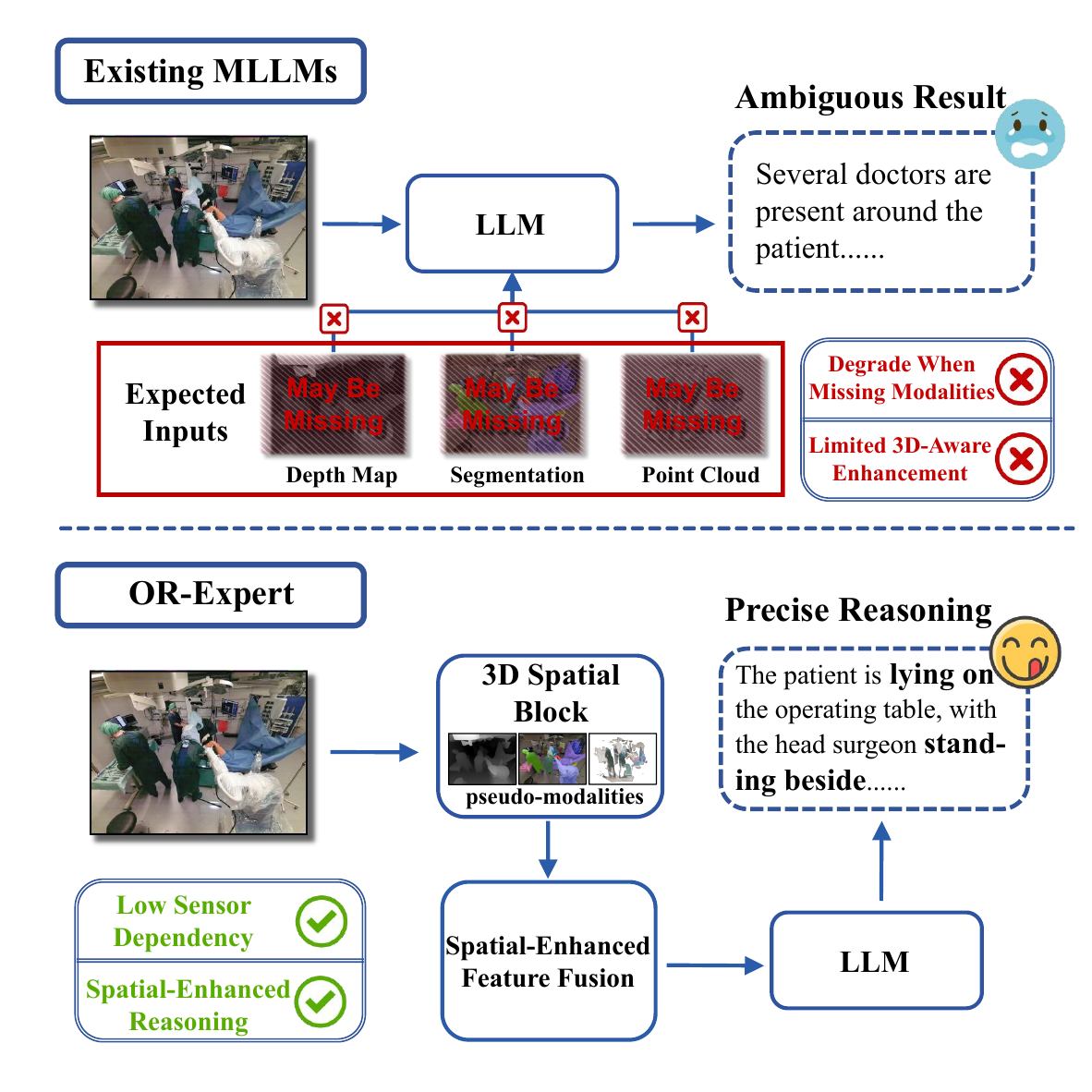}
    \caption{Comparison of existing multimodal LLMs and OR-Expert. OR-Expert reconstructs spatial cues from RGB images for 3D reasoning with RGB-only inference.}
    \label{fig:motivation}
\end{figure}

Recent vision-language models have substantially advanced \linebreak image-text alignment, visual question answering, and grounded language generation. For example, LLaVA~\cite{liu2023llava} connects a CLIP-based vision encoder with a large language model to support image-conditioned dialogue and general visual reasoning. Nevertheless, conventional vision-language models mainly operate on appearance features extracted from 2D images and lack explicit representations of geometric structure. Their spatial predictions therefore tend to rely on implicit visual correlations, which are insufficient for fine-grained reasoning about relative position, distance, orientation, and object interaction in crowded OR scenes.

Recent 3D-aware multimodal large language models have attempted to address this limitation by incorporating geometric information into visual-language representations. LLaVA-3D~\cite{zhu2024llava3d}, for instance, introduces 3D positional embeddings into a 2D MLLM framework to strengthen spatial understanding. Other approaches exploit multi-view images, RGB-D observations, or point-cloud inputs to provide explicit geometric cues. Although these modalities improve spatial reasoning, they are difficult to obtain consistently in clinical environments. Multi-view acquisition requires synchronized cameras and calibration, while external depth and point-cloud sensors increase deployment cost and system complexity. Moreover, geometry alone may be insufficient for distinguishing semantically similar or heavily occluded entities. Structured semantic priors, such as panoptic segmentation, are also important for separating medical staff, patients, instruments, and environmental objects.

These limitations are particularly consequential in OR environments, where ambiguous spatial interpretations may affect downstream scene understanding and decision support. Addressing them requires a model that can recover structured geometric and semantic cues from standard RGB observations while avoiding dependence on additional sensing modalities at inference time~\cite{v2razman2026systematic,10638254,bochkovskii2025depthprosharpmonocular}. As illustrated in Figure~\ref{fig:motivation}, the central objective is not to assume that general vision-language models cannot infer any spatial information from RGB images, but to provide explicit and structured 3D-aware representations for operating-room reasoning under modality-limited conditions.

A multimodal large language model is particularly suitable for this setting because it provides open-vocabulary understanding and natural-language interaction beyond predefined task labels. In contrast to task-specific models that typically produce fixed classes or relation categories, an MLLM can support diverse OR queries, spatial question answering, scene descriptions, scene graph generation, and extension to downstream clinical interfaces. However, effectively exploiting an MLLM requires spatial information to be represented as structured tokens that can participate in joint semantic, geometric, and textual reasoning, rather than being introduced as loosely concatenated auxiliary features.

To this end, we propose \textbf{OR-Expert}, a large vision-language framework for operating-room 3D spatial reasoning with RGB-only inference. During inference, OR-Expert requires only an RGB image as external input and internally generates three complementary pseudo-modalities: a depth map, a panoptic segmentation map, and a depth-derived point cloud. These representations encode relative geometry, semantic regions, and explicit 3D structure, respectively. A Spatial-Enhanced Feature Fusion Block projects the RGB and pseudo-modality features into a shared token space and integrates them with textual tokens for unified MLLM reasoning. This design enables OR-Expert to jointly model semantic entities and geometric relationships without requiring external depth, segmentation, or point-cloud sensor inputs at inference time.

Our main contributions are summarized as follows:

\begin{itemize}
    \item \textbf{OR-Expert Framework}: We introduce OR-Expert, a large vision-language framework for 3D spatial reasoning in \linebreak operating-room environments. It targets the practical setting of RGB-only inference by internally recovering structured geometric and semantic cues from monocular images.

    \item \textbf{Spatial-Enhanced Feature Fusion}: We develop a token-level fusion module that converts RGB-derived depth, panoptic segmentation, and point-cloud representations into structured spatial tokens and aligns them within a shared reasoning space. This design supports joint semantic, geometric, and textual reasoning in complex OR scenes.

    \item \textbf{Comprehensive Evaluation}: Experiments on multiple \linebreak operating-room benchmarks demonstrate state-of-the-art performance in spatial reasoning and scene graph generation. Results on unseen surgical scenes and cross-scene settings further validate the generalization capability of OR-Expert, while ablation studies confirm the complementary contributions of the pseudo-modalities and fusion design.
\end{itemize}
\section{Related Works}

\begin{figure*}[!t]
    \centering
    \includegraphics[width=\linewidth]{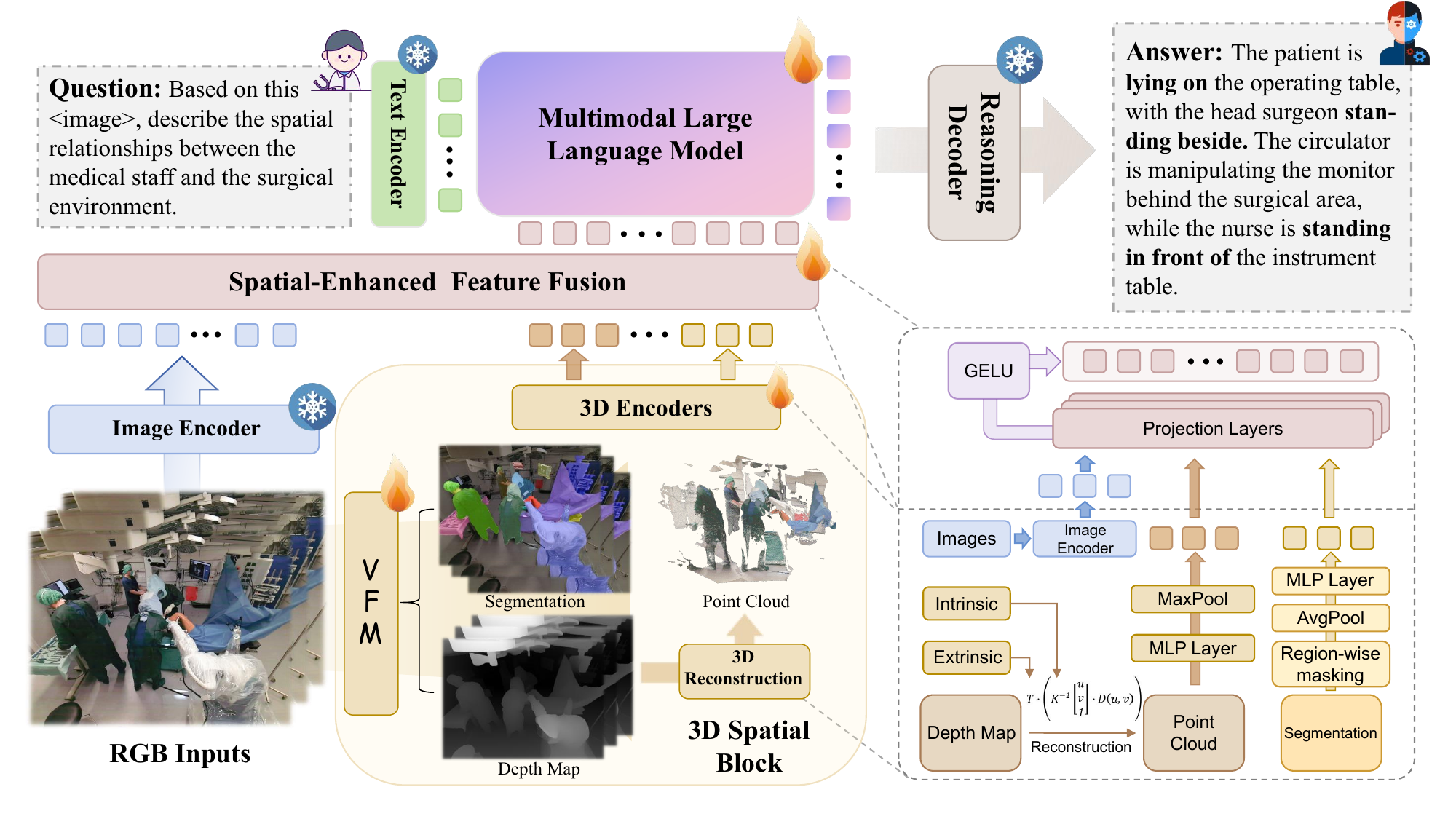}
    \caption{Architecture of OR-Expert. Given an RGB image, the 3D Spatial Block internally generates depth, panoptic segmentation, and point-cloud cues. These pseudo-modalities are encoded and projected into a shared token space, where they are fused with RGB and textual representations for spatial reasoning. OR-Expert therefore requires only RGB images as external input at inference time.}
    \label{fig:Mainfigure}
\end{figure*}

\subsection{Spatial Reasoning in Vision-Language Models}

Recent vision-language models have substantially advanced image-grounded understanding by connecting pretrained visual encoders with large language models. Representative systems, including Flamingo~\cite{Alayrac2022Flamingo,t2awadalla2023openflamingoopensourceframeworktraining}, BLIP~\cite{li2023blip,t1dai2023instructblipgeneralpurposevisionlanguagemodels}, and LLaVA~\cite{liu2023llava}, demonstrate strong performance in image captioning, visual question answering, and multimodal dialogue. More recent models, such as SpatialLLM~\cite{SpatialLLM}, Spatial-MLLM~\cite{SpatialMLLM}, and SpatialVLM~\cite{SpatialVLM}, further improve spatial understanding through spatially oriented instruction tuning and structured supervision. Nevertheless, these models predominantly encode appearance information from RGB images and often represent spatial relationships only implicitly. Their performance therefore remains limited when reasoning requires explicit geometric structure, relative depth, or fine-grained object relationships.

To strengthen geometry-aware reasoning, several multimodal large language models introduce explicit 3D priors. LLaVA-3D~\cite{zhu2024llava3d} augments visual tokens with 3D positional information derived from geometric observations. 3D-VisTA~\cite{zhu2023vista} learns joint 2D--3D representations from RGB-D data, while 3D-LLaVA~\cite{deng20253dllava} combines point-cloud prompts with a 3D-aware adapter for object-centric reasoning. VLM-3R~\cite{VLM-3R} incorporates instruction-aligned 3D reconstruction to improve reasoning from RGB observations. G$^2$VLM~\cite{G2} further unifies 3D reconstruction and geometry-grounded language reasoning within a shared framework. These studies demonstrate that explicitly grounding language representations in geometry can substantially improve spatial understanding.

However, many existing 3D-aware models depend on RGB-D observations, point clouds, LiDAR, or multi-view inputs during training or inference. Such modalities are difficult to acquire consistently in operating rooms because they require additional sensors, synchronization, and calibration. OR-Expert instead targets a modality-constrained clinical setting with RGB-only inference. It internally derives complementary geometric and semantic pseudo-modalities from a monocular RGB image and integrates them as structured spatial tokens, eliminating the need for external depth or point-cloud sensor inputs at inference time.

\subsection{Medical Scene Understanding}

Operating-room scene understanding has progressed with the development of specialized multimodal datasets. The 4D-OR dataset~\cite{ozsoy2022_4D_OR} provides synchronized multi-view RGB-D recordings, 3D point clouds, and scene-graph annotations describing surgical staff, instruments, and their interactions. MM-OR~\cite{ozsoy2024mmor} further expands this setting with synchronized RGB-D video, audio, robot logs, tracking information, panoptic segmentation, and semantic graph annotations. These resources support fine-grained modeling of surgical workflows and spatial relationships.

Based on these datasets, methods such as MM2SG~\cite{ozsoy2024mmor}, ORacle~\cite{ozsoy2024oracle}, LABRAD-OR~\cite{LABRAD-OR}, and TriTemp-OR~\cite{TriTemp-OR} model surgical entities and generate scene graphs or spatial triplets~\cite{v3chopra2025medmoe,v4lu2025integrating,v5chen2025adapting}. S$^2$Former-OR~\cite{S2} further introduces a single-stage bi-modal transformer for operating-room scene graph generation. Although these approaches achieve effective structured scene understanding, they commonly exploit synchronized multi-view observations, depth information, or other specialized sensor modalities. This dependence increases acquisition and deployment costs and limits applicability in operating rooms equipped only with standard monocular cameras~\cite{v8xin2025med3dvlm,v9lin2025tamingvisionlanguagemodelsmedical,v10yan2017perspectivetransformernetslearning,zhang2026unihmunifieddexteroushand}.

Recent visual foundation models provide an alternative means of recovering structured information from RGB images. Segment Anything~\cite{sam} offers transferable segmentation representations, while Depth Anything V2~\cite{depth_anything_v2} supports monocular depth estimation across diverse visual domains. Their medical adaptations have further improved robustness to blur, specular reflections, occlusions, and domain-specific anatomical structures in surgical imagery~\cite{lou2024surgicaldepthanythingdepth,kamtam2025surgisam2finetuningfoundationalmodel,t3yue2023surgicalsamefficientclasspromptable,t4MedSAM,zhang2025openhoiopenworldhandobjectinteraction}. OR-Expert builds on these advances by adapting pretrained visual modules to generate depth, panoptic segmentation, and point-cloud cues from RGB images. Unlike task-specific perception models, these representations are projected into a shared MLLM token space to support joint semantic, geometric, and textual reasoning for operating-room spatial understanding.
\section{Method}

\subsection{Overview}

As illustrated in Figure~\ref{fig:Mainfigure}, OR-Expert performs
operating-room spatial reasoning with RGB-only inference, where RGB
is the only scene-dependent sensing modality. The camera intrinsics
and extrinsics used for geometric back-projection are fixed calibration
parameters and do not provide scene-specific depth, segmentation, or
point-cloud observations. Given an RGB image
$I\in\mathbb{R}^{3\times H\times W}$, the 3D Spatial Block internally
generates three complementary pseudo-modalities: a depth map, a
panoptic segmentation map, and a depth-derived point cloud. These
representations provide relative geometry, semantic regions, and
explicit 3D structure, respectively. Modality-specific encoders and
projection heads then transform the RGB and pseudo-modality features
into a shared token space. The resulting spatial tokens are combined
with prompt tokens and processed by the LLM backbone of the MLLM
for joint semantic, geometric, and textual reasoning.

During training, depth and segmentation annotations supervise the
corresponding perception modules. At inference time, no externally
captured depth, segmentation, or point-cloud inputs are required; all
scene-dependent spatial cues are internally inferred from the RGB
image. The fixed camera calibration parameters are used only to map
the predicted depth into a common 3D coordinate system.

\subsection{3D Spatial Block}

The 3D Spatial Block extracts structured geometric and semantic
information from the input image. It contains a visual feature
modulator (VFM), which predicts depth and panoptic segmentation,
and a point-cloud reconstructor, which converts the predicted depth
into an explicit 3D representation.

\subsubsection{Depth Estimation}

Inspired by Depth Anything V2~\cite{depth_anything_v2}, the depth
branch predicts a dense depth map
$D\in\mathbb{R}^{H\times W}$:

\begin{equation}
D=\mathcal{F}_{d}(I),
\end{equation}

where $\mathcal{F}_{d}$ consists of a transformer-based encoder and
a lightweight upsampling decoder. The module is optimized using a
combination of pixel-wise and gradient consistency losses:

\begin{equation}
\mathcal{L}_{\mathrm{depth}}
=
\lambda_{\ell_1}\lVert D-D^{*}\rVert_1
+
\lambda_{\mathrm{grad}}
\lVert\nabla D-\nabla D^{*}\rVert_1,
\end{equation}

where $D^{*}$ denotes the ground-truth depth available during
training. The first term encourages depth accuracy, while the
gradient term preserves object boundaries and local geometric
continuity. The predicted depth is further encoded as

\begin{equation}
F_{\mathrm{depth}}=\mathcal{E}_{\mathrm{depth}}(D),
\end{equation}

providing structured cues about relative distance, surface layout,
and object contours.

\subsubsection{Panoptic Segmentation}

The segmentation branch is initialized from a pretrained segmentation
foundation model~\cite{sam} and adapted to operating-room imagery.
Given $I$, it predicts a panoptic segmentation map:

\begin{equation}
S=\mathcal{F}_{\mathrm{seg}}(I)
\in\{1,\ldots,K\}^{H\times W},
\end{equation}

where $K$ denotes the number of semantic categories. The branch is
trained using cross-entropy and Dice losses:

\begin{equation}
\begin{split}
\mathcal{L}_{\mathrm{seg}}
={}&
-\sum_{u,v,k}
y^{*}_{u,v,k}\log\hat{p}_{u,v,k} \\
&+
\lambda_{\mathrm{Dice}}
\sum_{k}
\left(
1-\mathrm{Dice}_{k}(\hat{p},y^{*})
\right),
\end{split}
\end{equation}

where $\hat{p}_{u,v,k}$ is the predicted probability of category
$k$ at pixel $(u,v)$ and $y^{*}$ is the one-hot ground-truth
annotation.

To preserve region-level semantics, features are pooled separately
within each predicted region:

\begin{equation}
F_{\mathrm{seg}}
=
\left[f_r\right]_{r=1}^{R},
\qquad
f_r
=
\phi\!\left(
\mathrm{Pool}
\left(
F_{\mathrm{vfm}}\odot M_r
\right)
\right),
\end{equation}

where $M_r$ is the mask of region $r$, $F_{\mathrm{vfm}}$ denotes
the intermediate VFM feature map, and $\phi(\cdot)$ is a lightweight
MLP. The resulting region tokens retain both semantic category
information and localized visual context.

\subsubsection{Point-Cloud Reconstruction}

To provide explicit 3D geometry, we reconstruct a partial point cloud
from the predicted depth map. Let $K$ and $T$ denote the pre-calibrated
camera intrinsics and extrinsics, respectively. These parameters are
fixed for each camera and provide no scene-dependent geometric
observations. They are used only for deterministic back-projection
of the predicted depth. The pixel $(u,v)$ is back-projected as

\begin{equation}
\mathbf{p}_{u,v}
=
T
\left(
D(u,v)
K^{-1}
\begin{bmatrix}
u & v & 1
\end{bmatrix}^{\top}
\right),
\end{equation}

where $T$ maps the reconstructed point from the camera coordinate
system to the reference coordinate system. Collecting the valid
back-projected pixels produces the point set

\begin{equation}
\mathcal{P}
=
\left\{
\mathbf{p}_{u,v}
\mid
(u,v)\in\Omega
\right\},
\end{equation}

where $\Omega$ denotes the valid image domain. The reconstructed
point cloud makes the relative arrangement of visible surfaces
explicit and complements the implicit geometry encoded by the
depth map. Thus, the point cloud is derived entirely from the RGB
prediction pipeline rather than from an externally captured spatial
modality.

A permutation-invariant point-cloud encoder~\cite{PointNet++}
converts $\mathcal{P}$ into geometric features:

\begin{equation}
F_{\mathrm{pc}}
=
\mathcal{E}_{\mathrm{pc}}(\mathcal{P}),
\end{equation}

where $\mathcal{E}_{\mathrm{pc}}$ applies shared point-wise
transformations and symmetric aggregation. The resulting
representation captures relative distances, object extents, and
global scene layout.

\begin{table*}[t]
\centering
\setlength{\tabcolsep}{3.6pt}
\caption{
Performance comparison on seen OR scenes.
AbsRel, mIoU, and CD evaluate geometric accuracy, while R-L, M, C,
and EM evaluate spatial reasoning.
``$*$'' denotes adaptation with a lightweight task head, and ``--''
denotes unavailable or unreported results.
}
\label{tab:maintable_seen}

\begin{tabular}{l|c|ccc|cccc}
\toprule
\textbf{Model}
& \textbf{Input}
& \textbf{AbsRel$\downarrow$}
& \textbf{mIoU$\uparrow$}
& \textbf{CD$\downarrow$}
& \textbf{R-L$\uparrow$}
& \textbf{M$\uparrow$}
& \textbf{C$\uparrow$}
& \textbf{EM$\uparrow$} \\
\midrule

\multicolumn{9}{l}{\textit{\textbf{2D LLMs}}} \\

GPT-4V
& RGB
& -- & -- & --
& 41.7 & 33.1 & 56.2 & 30.6 \\

LLaVA-OV$^*$~\cite{llavaov}
& RGB
& 0.275$^*$ & 48.6$^*$ & 3.12$^*$
& 46.6 & 38.8 & 60.1 & 28.7 \\

\midrule
\multicolumn{9}{l}{\textit{\textbf{OR-Domain Models}}} \\

ORacle$^*$~\cite{ozsoy2024oracle}
& RGB
& 0.249$^*$ & 50.9$^*$ & 2.86$^*$
& 53.1 & 44.0 & 87.1 & 55.8 \\

MM2SG$^*$~\cite{ozsoy2024mmor}
& ALL
& 0.238$^*$ & 52.7$^*$ & 2.73$^*$
& 58.2 & 51.0 & 91.7 & 61.5 \\

S$^2$Former-OR~\cite{S2}
& RGB
& -- & -- & --
& 56.8 & -- & -- & 63.1 \\

\midrule
\multicolumn{9}{l}{\textit{\textbf{3D-Aware LLMs}}} \\

3D-LLM~\cite{3dllm}
& RGB+PC
& -- & 50.4 & --
& 47.0 & 41.1 & 73.3 & 56.5 \\

LEO~\cite{LEO}
& RGB+PC
& -- & 52.7 & --
& 56.9 & 49.8 & 88.5 & 61.2 \\

Chat-Scene~\cite{Chat-scene}
& RGB+PC
& -- & 51.6 & --
& 55.6 & 52.2 & 88.8 & 57.7 \\

3D-LLaVA~\cite{deng20253dllava}
& PC
& -- & 53.3 & --
& 56.3 & 50.0 & 90.9 & 50.2 \\

LLaVA-3D$^*$~\cite{zhu2024llava3d}
& RGB
& 0.216 & 51.2$^*$ & 2.64$^*$
& 58.9 & 50.5 & 91.4 & 61.8 \\

G$^2$VLM~\cite{G2}
& RGB
& 0.207 & 53.1 & 2.49
& \textbf{61.5} & 56.7 & 94.8 & 66.9 \\

\midrule
\textbf{OR-Expert (Ours)}
& \textbf{RGB}
& \textbf{0.198}
& \textbf{53.7}
& \textbf{2.41}
& 61.2
& \textbf{58.6}
& \textbf{96.4}
& \textbf{67.8} \\

\bottomrule
\end{tabular}
\end{table*}

\subsection{Spatial-Enhanced Feature Fusion}

The Spatial-Enhanced Feature Fusion Block aligns heterogeneous RGB,
depth, segmentation, and point-cloud representations within the token
space of the MLLM. Rather than directly concatenating raw modality
features, OR-Expert first applies modality-specific encoders and
learnable projection heads, allowing each representation to be mapped
into a shared semantic and geometric space.

The RGB image tokens are obtained using a pretrained CLIP encoder:

\begin{equation}
T_{\mathrm{img}}
=
\mathcal{P}_{\mathrm{img}}
\left(
\mathcal{E}_{\mathrm{CLIP}}(I)
\right),
\end{equation}

where $\mathcal{E}_{\mathrm{CLIP}}$ extracts patch-level visual
features and $\mathcal{P}_{\mathrm{img}}$ maps them to the MLLM
token dimension $d_{\mathrm{token}}$.

The pseudo-modality representations are projected in the same manner:

\begin{equation}
\begin{aligned}
T_{\mathrm{depth}}
&=
\mathcal{P}_{\mathrm{depth}}
(F_{\mathrm{depth}}),\\
T_{\mathrm{seg}}
&=
\mathcal{P}_{\mathrm{seg}}
(F_{\mathrm{seg}}),\\
T_{\mathrm{pc}}
&=
\mathcal{P}_{\mathrm{pc}}
(F_{\mathrm{pc}}).
\end{aligned}
\end{equation}

Each projection head contains two linear layers separated by a GELU
activation. The projected tokens therefore share the same
dimensionality while retaining modality-specific geometric or
semantic information.

The final multimodal sequence is constructed as

\begin{equation}
T_{\mathrm{input}}
=
\left[
T_{\mathrm{img}};
T_{\mathrm{seg}};
T_{\mathrm{pc}};
T_{\mathrm{prompt}}
\right],
\end{equation}

where $T_{\mathrm{prompt}}$ denotes the textual instruction tokens.
The fused sequence is processed by the LLM backbone:

\begin{equation}
\label{eq:llm_fusion}
Y
=
\mathcal{F}_{\mathrm{LLM}}
\left(
T_{\mathrm{input}};
\theta_{\mathrm{LLM}}
\right),
\end{equation}

where $Y$ is the generated spatial description, answer, or structured
scene interpretation.

This token-level design enables self-attention to establish
cross-modal correspondences between visual appearance, semantic
regions, depth structure, and reconstructed geometry. For example,
RGB tokens describing an operating table can attend to segmentation
tokens identifying the patient and staff, while depth and point-cloud
tokens provide their relative positions. Consequently, OR-Expert
performs compositional spatial reasoning over jointly aligned
semantic, geometric, and textual evidence rather than relying on
implicit RGB correlations alone.
\section{Experiments}
\begin{table*}[t]
\centering
\setlength{\tabcolsep}{3.6pt}
\caption{
Zero-shot generalization results on unseen OR scenes.
AbsRel, mIoU, and CD evaluate geometric accuracy, while R-L, M, C,
and EM evaluate spatial reasoning.
``--'' denotes unavailable or unreported results.
}
\label{tab:maintable_unseen}

\begin{tabular}{l|c|ccc|cccc}
\toprule
\textbf{Model}
& \textbf{Input}
& \textbf{AbsRel$\downarrow$}
& \textbf{mIoU$\uparrow$}
& \textbf{CD$\downarrow$}
& \textbf{R-L$\uparrow$}
& \textbf{M$\uparrow$}
& \textbf{C$\uparrow$}
& \textbf{EM$\uparrow$} \\
\midrule

\multicolumn{9}{l}{\textit{\textbf{2D LLMs}}} \\

GPT-4V
& RGB
& -- & -- & --
& 35.8 & 28.5 & 48.3 & 26.4 \\

LLaVA-OV~\cite{llavaov}
& RGB
& -- & -- & --
& 37.2 & 29.8 & 47.5 & 21.6 \\

\midrule
\multicolumn{9}{l}{\textit{\textbf{OR-Domain Models}}} \\

S$^2$Former-OR~\cite{S2}
& RGB
& -- & -- & --
& 44.7 & 39.2 & 72.5 & 48.9 \\

\midrule
\multicolumn{9}{l}{\textit{\textbf{3D-Aware LLMs}}} \\

3D-LLM~\cite{3dllm}
& RGB+PC
& -- & 38.3 & --
& 37.7 & 31.9 & 59.3 & 44.2 \\

LEO~\cite{LEO}
& RGB+PC
& -- & 40.1 & --
& 43.6 & 37.8 & 70.6 & 48.4 \\

Chat-Scene~\cite{Chat-scene}
& RGB+PC
& -- & 39.1 & --
& 42.6 & 38.9 & 71.3 & 45.2 \\

3D-LLaVA~\cite{deng20253dllava}
& PC
& -- & 40.8 & --
& 43.4 & 39.5 & 73.4 & 40.9 \\

LLaVA-3D~\cite{zhu2024llava3d}
& RGB
& 0.203 & -- & --
& 45.1 & 39.9 & 74.0 & 47.6 \\

G$^2$VLM~\cite{G2}
& RGB
& 0.196 & 41.8 & 2.34
& 47.6 & 42.8 & 76.5 & 50.8 \\

\midrule
\textbf{OR-Expert (Ours)}
& \textbf{RGB}
& \textbf{0.185}
& \textbf{42.6}
& \textbf{2.19}
& \textbf{48.8}
& \textbf{44.5}
& \textbf{78.2}
& \textbf{52.0} \\

\bottomrule
\end{tabular}
\end{table*}
\begin{table}[!t]
\centering
\caption{Scene graph generation results on seen OR scenes.}
\label{tab:scene_graph_metrics}
\begin{tabular}{@{}lccc@{}}
\toprule
\textbf{Model}
& \textbf{P(\%)$\uparrow$}
& \textbf{R(\%)$\uparrow$}
& \textbf{F1(\%)$\uparrow$} \\
\midrule
GPT-4V
& 52.6 & 41.9 & 46.6 \\

MM2SG~\cite{ozsoy2024mmor}
& 45.3 & 39.4 & 42.1 \\

ORacle~\cite{ozsoy2024oracle}
& 49.8 & 42.1 & 45.6 \\

S$^2$Former-OR~\cite{S2}
& 54.2 & 47.0 & 50.3 \\

LLaVA-3D~\cite{zhu2024llava3d}
& 53.2 & 46.7 & 49.7 \\

G$^2$VLM~\cite{G2}
& 55.1 & 47.8 & 51.2 \\

\textbf{OR-Expert}
& \textbf{55.4}
& \textbf{48.3}
& \textbf{51.6} \\
\bottomrule
\end{tabular}
\end{table}
\begin{table}[t]
\centering
\caption{Zero-shot scene graph generation on unseen OR scenes.}
\label{tab:scene_graph_zeroshot}
\begin{tabular}{@{}lccc@{}}
\toprule
\textbf{Model}
& \textbf{P(\%)$\uparrow$}
& \textbf{R(\%)$\uparrow$}
& \textbf{F1(\%)$\uparrow$} \\
\midrule
GPT-4V
& 42.2 & 41.5 & 41.8 \\

LLaVA-3D
& 47.8 & 41.6 & 44.5 \\

\textbf{OR-Expert}
& \textbf{49.8}
& \textbf{42.7}
& \textbf{46.0} \\
\bottomrule
\end{tabular}
\end{table}
\subsection{Datasets}

We conduct experiments on two operating-room datasets, MM-OR~\cite{ozsoy2024mmor} and 4D-OR~\cite{ozsoy2022_4D_OR}. Both datasets record total knee replacement procedures using synchronized multi-view RGB-D cameras and provide camera calibration parameters, depth observations, panoptic annotations, and expert-annotated scene graphs.

GPT assistance is used for keyframe selection and for augmenting sparse textual descriptions and implicit spatial relations during training. These descriptions serve only as auxiliary supervision, while the original expert annotations are preserved as the ground truth for evaluation.

After preprocessing, the combined dataset contains 46,742 samples with RGB images, depth maps, panoptic masks, scene graphs, and textual descriptions. To prevent adjacent-frame leakage, we split the data by operating-room scenes rather than individual frames, following a 60/20/20 ratio for training, validation, and testing.

\subsection{Training Strategy}

OR-Expert adopts Vicuna-v1.5-7B as the LLM backbone and is implemented in PyTorch based on the LLaVA training framework. The visual modules are initialized from pretrained models and adapted through two-stage training.

\textbf{Stage 1: Visual and Spatial Alignment.}
We freeze the LLM and optimize the CLIP ViT-L/14 encoder, depth and segmentation backbones, point-cloud encoder, and modality-specific projection heads. RGB and pseudo-modality representations are aligned within a shared token space using multimodal alignment and next-token objectives.

\textbf{Stage 2: Domain Instruction Tuning.}
We freeze the visual modules and fine-tune Vicuna using OR-specific question answering, scene-graph descriptions, and spatial reasoning instructions. GPT-augmented descriptions are used only as auxiliary training signals, and the same prompt templates are used across datasets.

\textbf{Implementation and Complexity.}
Training is conducted on eight NVIDIA A100 GPUs using AdamW and an effective batch size of 128. Stage 1 uses learning rates of $1\times10^{-3}$ for newly introduced layers and $5\times10^{-4}$ for pretrained visual modules, while Stage 2 uses $2\times10^{-5}$ for Vicuna. Both stages are trained for 10 epochs, resulting in 20 epochs in total. For a fused sequence of $N$ tokens with hidden dimension $d$, the dominant LLM attention cost is $\mathcal{O}(N^{2}d)$, while the modality-specific projection cost grows linearly with $N$. The overall computation is therefore dominated by the 7B language backbone and pretrained visual encoders rather than the lightweight projection heads.

\subsection{Evaluation Metrics}

We evaluate geometric reconstruction, language-based spatial reasoning, and structured scene understanding. Absolute Relative Error (AbsRel), Mean Intersection over Union (mIoU), and Chamfer Distance (CD) assess depth estimation, panoptic segmentation, and point-cloud reconstruction, respectively. ROUGE-L (R-L), METEOR (M), CIDEr (C), Exact Match (EM), and Spatial Relation Accuracy (Rel-Acc) evaluate generated spatial descriptions. For scene graph generation, we report Precision (P), Recall (R), and F1-score.

\begin{table*}[!t]
\centering
\setlength{\tabcolsep}{3.6pt}
\caption{
Cross-scene generalization results on ScanNet~\cite{scannet}.
``$*$'' denotes adaptation with a lightweight task head, and ``--''
denotes unavailable or unreported results.
}
\label{tab:cross-scene}

\begin{tabular}{l|c|ccc|cccc}
\toprule
\textbf{Model}
& \textbf{Input}
& \textbf{AbsRel$\downarrow$}
& \textbf{mIoU$\uparrow$}
& \textbf{CD$\downarrow$}
& \textbf{R-L$\uparrow$}
& \textbf{M$\uparrow$}
& \textbf{C$\uparrow$}
& \textbf{EM$\uparrow$} \\
\midrule

\multicolumn{9}{l}{\textit{\textbf{2D LLMs}}} \\

GPT-4V
& RGB
& -- & -- & --
& 33.1 & 26.4 & 45.2 & 21.6 \\

LLaVA-OV$^*$~\cite{llavaov}
& RGB
& 0.334$^*$ & 38.9$^*$ & 3.71$^*$
& 35.6 & 28.3 & 47.9 & 19.8 \\

\midrule
\multicolumn{9}{l}{\textit{\textbf{OR-Domain Models}}} \\

ORacle$^*$~\cite{ozsoy2024oracle}
& RGB
& 0.302$^*$ & 41.6$^*$ & 3.44$^*$
& 38.7 & 31.9 & 58.1 & 33.4 \\

MM2SG$^*$~\cite{ozsoy2024mmor}
& ALL
& 0.287$^*$ & 42.8$^*$ & 3.31$^*$
& 41.9 & 34.8 & 61.7 & 36.5 \\

S$^2$Former-OR~\cite{S2}
& RGB
& -- & -- & --
& 44.2 & 38.1 & 64.9 & 39.1 \\

\midrule
\multicolumn{9}{l}{\textit{\textbf{3D-Aware LLMs}}} \\

3D-LLM~\cite{3dllm}
& RGB+PC
& -- & 40.8 & --
& 38.6 & 32.4 & 54.9 & 34.7 \\

LEO~\cite{LEO}
& RGB+PC
& -- & 43.7 & --
& 43.9 & 38.5 & 63.2 & 38.6 \\

Chat-Scene~\cite{Chat-scene}
& RGB+PC
& -- & 42.9 & --
& 43.2 & 39.1 & 64.0 & 37.4 \\

3D-LLaVA~\cite{deng20253dllava}
& PC
& -- & 44.1 & --
& 44.7 & 39.0 & 66.3 & 35.2 \\

LLaVA-3D$^*$~\cite{zhu2024llava3d}
& RGB
& 0.259 & 43.6$^*$ & 3.18$^*$
& 45.4 & 40.2 & 67.1 & 39.7 \\

G$^2$VLM~\cite{G2}
& RGB
& 0.251 & 45.0 & 3.08
& 47.2 & 41.5 & 69.6 & 41.4 \\

\midrule
\textbf{OR-Expert (Ours)}
& \textbf{RGB}
& \textbf{0.244}
& \textbf{45.8}
& \textbf{2.97}
& \textbf{48.1}
& \textbf{42.6}
& \textbf{71.3}
& \textbf{42.8} \\

\bottomrule
\end{tabular}
\end{table*}
\begin{table}[t]
\centering
\caption{Expert preference rates in the user study.}
\label{tab:user_study}
\footnotesize
\renewcommand{\arraystretch}{1.05}
\setlength{\tabcolsep}{4pt}

\begin{tabular*}{\columnwidth}{
@{\extracolsep{\fill}}
lccc
@{}
}
\toprule
Scene
& S$^2$Former-OR~\cite{S2}
& G$^2$VLM~\cite{G2}
& \textbf{OR-Expert} \\
\midrule
Seen OR
& 21.6\%
& 25.4\%
& \textbf{53.0\%} \\

Unseen OR
& 19.8\%
& 23.7\%
& \textbf{56.5\%} \\
\bottomrule
\end{tabular*}
\end{table}
\begin{figure}[t]
    \centering
    \includegraphics[width=0.9\linewidth]{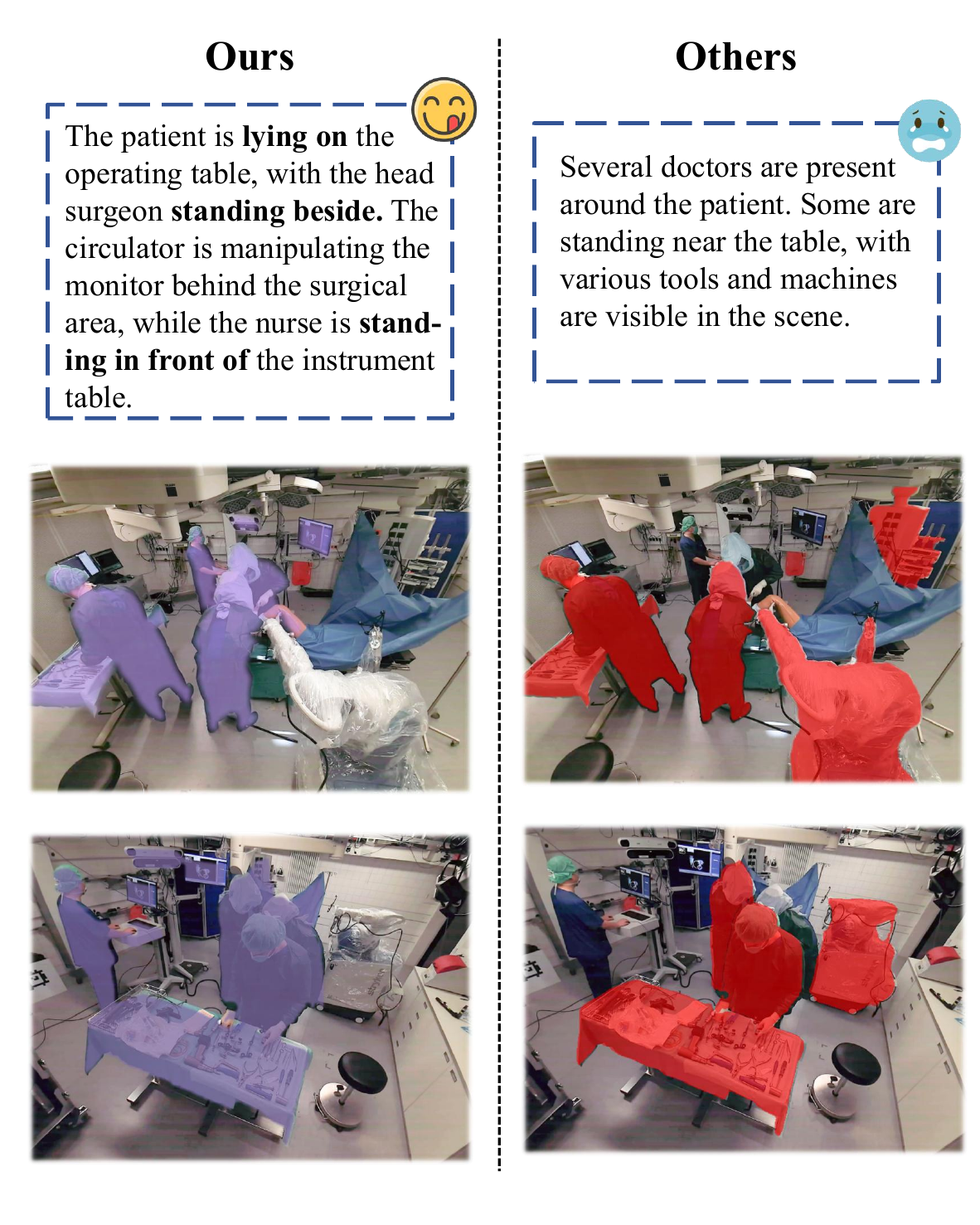}
    \caption{Qualitative comparison between OR-Expert and baseline models.}
    \Description{Comparison of spatial descriptions generated for an operating-room scene. OR-Expert describes detailed relationships among the patient, medical staff, monitor, and instrument table, whereas the baseline output provides a more generic description.}
    \label{fig:compare}
\end{figure}
\subsection{Main Results}

\textbf{Seen OR Scenes.}
For fair comparison, all trainable baselines use the same training data, scene-level split, prompt templates, and supervision as OR-Expert. Models marked with ``$*$'' are equipped with lightweight task heads only to produce outputs compatible with the common evaluation protocol. GPT-4V is evaluated through its API using dataset examples as in-context demonstrations.

As shown in Table~\ref{tab_maintable_seen}, OR-Expert achieves the strongest overall performance across geometric reconstruction and spatial reasoning metrics. Compared with conventional VLMs, OR-domain models, and geometry-aware MLLMs, it produces more accurate geometric representations and more spatially consistent language outputs. 

\textbf{Unseen OR Scenes.}
For zero-shot evaluation, OR-Expert is trained on the generic
\textit{LLaVA-3D-Instruct-86K} dataset~\cite{zhu2024llava3d} without using OR-specific training data. Other open-source methods use their publicly released pretrained weights and are directly evaluated on the OR datasets. GPT-4V receives no OR-specific in-context examples in this setting.

Although all methods experience degradation under zero-shot transfer, OR-Expert maintains the strongest overall results. Its advantage over S$^2$Former-OR~\cite{S2} and G$^2$VLM~\cite{G2} indicates that the proposed spatial representations transfer effectively to unseen surgical scenes.

\subsection{Scene Graph Generation}

We further evaluate OR-Expert on scene graph generation (SGG), which requires the identification of semantic entities and prediction of their pairwise spatial relationships~\cite{xu2017scenegraphgenerationiterative}. The original scene graph annotations from MM-OR and 4D-OR are used without modification.

All methods receive RGB images as the external input at inference time. OR-Expert internally derives depth, segmentation, and point-cloud cues without accessing ground-truth spatial modalities.

As shown in Table~\ref{tab_scene_graph_metrics}, OR-Expert achieves the best structured scene understanding performance. Its advantage over OR-specific and geometry-aware baselines demonstrates that the proposed spatial enhancement benefits not only free-form description generation but also explicit entity-relation prediction.

We additionally conduct zero-shot SGG evaluation on unseen OR scenes. OR-Expert and LLaVA-3D~\cite{zhu2024llava3d} are trained without OR-specific data, while GPT-4V is evaluated directly through its API without additional adaptation.

The results in Table~\ref{tab_scene_graph_zeroshot} show that OR-Expert maintains its advantage under zero-shot transfer, indicating that the learned representations remain effective for structured reasoning in unseen OR environments.

\begin{table*}[!ht]
\centering
\setlength{\tabcolsep}{4.5pt}
\caption{
Ablation results for pseudo-modalities, fusion strategies, LLM settings,
and spatial inputs. ``GT'' denotes ground-truth spatial inputs, and
``--'' denotes an unreported result.
}
\label{tab:ablation}

\begin{tabular}{l|ccc|ccc}
\toprule
\textbf{Model Variant}
& \textbf{R-L$\uparrow$}
& \textbf{EM$\uparrow$}
& \textbf{Rel-Acc$\uparrow$}
& \textbf{P$\uparrow$}
& \textbf{R$\uparrow$}
& \textbf{F1$\uparrow$} \\
\midrule

\multicolumn{7}{l}{\textit{\textbf{Pseudo-Modality Ablations}}} \\

w/o Point Cloud Reconstruction
& 58.9 & 64.6 & --
& 53.8 & 46.6 & 49.9 \\

w/o Panoptic Segmentation
& 58.2 & 61.2 & --
& 53.1 & 45.9 & 49.2 \\

w/o Depth Estimation
& 50.5 & 43.2 & --
& 48.6 & 42.8 & 45.5 \\

w/o Depth Est. \& Panoptic Seg.
& 39.5 & 37.8 & --
& 44.7 & 39.6 & 42.0 \\

\midrule
\multicolumn{7}{l}{\textit{\textbf{Fusion and Backbone Ablations}}} \\

Ours w/ Early Pixel Fusion
& 57.4 & 62.8 & 61.9
& 51.7 & 44.6 & 47.9 \\

Ours w/ Gated Fusion
& 58.7 & 64.7 & 63.4
& 53.1 & 46.1 & 49.4 \\

Ours w/ Direct Depth+PC+Seg Fusion
& 61.0 & 67.5 & 65.8
& \textbf{55.7} & 47.7 & 51.4 \\

Ours w/ Qwen2.5-7B-Instruct
& \textbf{61.7} & 67.3 & 65.0
& 55.0 & 48.1 & 51.3 \\

\midrule
\multicolumn{7}{l}{\textit{\textbf{Training and Input Ablations}}} \\

Ours w/ Frozen LLM
& 58.9 & 64.1 & 62.7
& 52.8 & 45.8 & 49.0 \\

Ours w/ GT Spatial Inputs
& 61.4 & \textbf{68.1} & \textbf{66.4}
& 55.6 & \textbf{48.5} & \textbf{51.8} \\

\midrule
\textbf{OR-Expert (Full)}
& 61.2
& 67.8
& 66.1
& 55.4
& 48.3
& 51.6 \\

\bottomrule
\end{tabular}
\end{table*}

\subsection{Cross-Scene Generalization}

To evaluate generalization beyond surgical environments, we test the models on ScanNet~\cite{scannet}, whose indoor scenes differ substantially from OR data in layout, appearance, and object categories.

OR-Expert is trained on the OR datasets and directly evaluated on ScanNet without additional fine-tuning. Other methods use their publicly released pretrained weights or original training configurations. Models marked with ``$*$'' are adapted using lightweight task heads for compatibility with the common evaluation protocol.

As reported in Table~\ref{tab_cross-scene}, OR-Expert demonstrates stronger robustness under cross-domain transfer, suggesting that its geometric and semantic representations remain effective under substantial scene and category shifts.

\subsection{User Study}

We conduct a user study on 500 test samples to evaluate the practical quality of the generated spatial descriptions. Medical-background experts compare the outputs of S$^2$Former-OR, G$^2$VLM, and OR-Expert and select the response that best matches the spatial relationships and clinical interpretation of each scene.

As shown in Table~\ref{tab_user_study}, OR-Expert receives the highest expert preference in both seen and unseen OR settings, providing complementary evidence that its descriptions are more spatially grounded and clinically coherent.

\subsection{Visualization}

We provide a qualitative case study comparing the spatial descriptions generated by OR-Expert and baseline methods on a representative surgical scene.

As illustrated in Figure~\ref{fig:compare}, baseline methods identify visible entities but provide generic descriptions. In contrast, OR-Expert explicitly describes their relative positions and interactions, demonstrating the benefit of combining semantic and geometric pseudo-modalities.

\subsection{Ablation Study}

We examine the contributions of the pseudo-modalities, fusion strategy, LLM backbone, fine-tuning configuration, and spatial inputs. The variants remove spatial branches, replace token-level fusion, change the LLM backbone, freeze the LLM, or replace predicted pseudo-modalities with ground-truth spatial inputs.

As shown in Table~\ref{tab_ablation}, depth, panoptic segmentation, and point-cloud representations provide complementary spatial information. Removing individual branches degrades both language reasoning and scene graph generation, while jointly removing depth and segmentation produces the largest decline.

Early Pixel Fusion and Gated Fusion underperform the proposed Spatial-Enhanced Feature Fusion, confirming the benefit of structured token-level alignment. The Direct Depth+PC+Seg Fusion variant remains competitive but provides a weaker overall balance.

Qwen2.5-7B-Instruct produces comparable results, indicating that the method is not restricted to Vicuna. Freezing the LLM reduces performance, demonstrating the importance of domain instruction tuning. Finally, ground-truth spatial inputs yield only limited improvements, suggesting that the RGB-derived pseudo-modalities already recover most of the geometric and semantic information required for OR reasoning.
\section{Conclusion}

We presented OR-Expert, a multimodal large language model for 3D spatial reasoning with RGB-only inference in operating-room environments. By internally generating depth, panoptic segmentation, and point-cloud cues and aligning them within a shared token space, OR-Expert jointly captures semantic and geometric information without requiring external spatial sensors at inference time. Extensive experiments on MM-OR and 4D-OR demonstrate strong performance in spatial reasoning and scene graph generation, while evaluations on unseen OR scenes and ScanNet further confirm the robustness and transferability of the learned spatial representations.

The results show that explicitly introducing structured RGB-derived pseudo-modalities can substantially strengthen spatial understanding under modality-constrained clinical settings. OR-Expert is intended as a perception and reasoning component rather than an autonomous clinical decision system. Future work will further investigate real-time efficiency, robustness under more diverse surgical conditions, and prospective validation toward practical integration into surgical assistance systems.

\newpage

\begin{acks}
The authors thank the anonymous reviewers for their constructive comments and acknowledge the creators of the MM-OR and 4D-OR datasets for making these valuable resources available. Shaoliang Peng provided overall academic supervision and played a central role in shaping the research direction, methodological design, and interpretation of the results. Zhenhao Zhang led project administration and coordinated the overall planning, implementation, and evaluation of this work.

This work was supported by the National Key R\&D Program of China under Grant Nos.~2023YFC3503400 and 2022YFC3400400; the NSFC--FDCT Joint Research Program under Grant No.~62361166662; the Innovative Research Group Project of Hunan Province under Grant No.~2024JJ1002; the Hunan Science and Technology Innovation Plan under Grant No.~2025ZYJ003; the Key R\&D Program of Hunan Province under Grant Nos.~2023GK2004, 2023SK2059, and 2023SK2060; the Top 10 Technical Key Project in Hunan Province under Grant No.~2023GK1010; and the Key Technologies R\&D Program of Guangdong Province under Grant No.~2023B1111030004 (to FFH). Additional support was provided by the State Key Laboratory of Chemo and Biosensing, the National Supercomputing Center in Changsha (\url{http://nscc.hnu.edu.cn/}), Peng Cheng Laboratory, the Key Laboratory of High-Performance Distributed Ledger Technology and Digital Finance (Ministry of Education), and the Hunan Research Center of the Basic Discipline for Cell Signaling.
\end{acks}

\bibliographystyle{ACM-Reference-Format}
\bibliography{reference}

\end{document}